\documentclass[conference]{IEEEtran}
\IEEEoverridecommandlockouts

\usepackage{cite}
\usepackage{amsmath,amssymb,amsfonts}
\usepackage{algorithmic}
\usepackage{textcomp}
\usepackage{xcolor}

\def\BibTeX{{\rm B\kern-.05em{\sc i\kern-.025em b}\kern-.08em
    T\kern-.1667em\lower.7ex\hbox{E}\kern-.125emX}}

\usepackage[T1]{fontenc}	
\usepackage{mathtools}
\usepackage{siunitx}
\usepackage{url}
\usepackage{bm}

\usepackage{graphicx}
\usepackage[font={small}]{caption} 
\usepackage{subcaption}
\usepackage{wrapfig}

\usepackage{hyperref}
\hypersetup{
    colorlinks=true,
    linkcolor=blue,
    filecolor=magenta,      
    urlcolor=blue,
}

\usepackage{tikz}     
\usepackage{pgfplots}
\usepackage{tikz-layers}

\usepackage{booktabs}
\usepackage{multirow}

\newcommand{\op}[1]{\operatorname{#1}}

\DeclarePairedDelimiterX{\infdivx}[2]{(}{)}{%
#1\;\delimsize\|\;#2%
}

\pgfplotsset{every axis/.append style={
		label style={font=\tiny},
		tick label style={font=\tiny},  
		/tikz/inner sep to outer sep/.style={inner sep=0pt, outer sep=.3333em},
		x tick label style=inner sep to outer sep,
		x label style=inner sep to outer sep,
		y label style=inner sep to outer sep,
		try min ticks=3,                            
		major tick length=0.10cm                    
	}
}
\pgfdeclarelayer{background}
\pgfdeclarelayer{foreground}
\pgfdeclarelayer{behindforeground}
\pgfsetlayers{background,main,behindforeground,foreground}
\pgfplotsset{
	layers/my layer set/.define layer set={
		background,
		main,
		behindforeground,
		foreground
	}{
	},
	set layers=my layer set,
}

\pgfplotsset{compat=newest}
\usetikzlibrary{decorations.pathmorphing}	
\usetikzlibrary{fit}	
\usetikzlibrary{pgfplots.groupplots}
\usetikzlibrary{external}
\renewcommand{\vec}[1]{\boldsymbol{#1}}
\newcommand{\mat}[1]{\boldsymbol{\mathrm #1}}

\begin{document}

\title{\vspace{0.5cm} A Variational Infinite Mixture for \\ Probabilistic Inverse Dynamics Learning}

\author{
    \IEEEauthorblockN{
        Hany Abdulsamad, \quad Peter Nickl, \quad Pascal Klink, \quad Jan Peters \thanks{This project received funding from the European Union’s Horizon 2020 research and innovation program under grant agreement No \#640554 (SKILLS4ROBOTS) and the DFG project PA3179/1-1 (ROBOLEAP).}
        }
    \IEEEauthorblockA{
        \textit{Intelligent Autonomous Systems}, 
        \textit{Technische Universität Darmstadt} \\
        }
}

\maketitle

\begin{abstract}
    Probabilistic regression techniques in control and robotics applications have to fulfill different criteria of data-driven adaptability, computational efficiency, scalability to high dimensions, and the capacity to deal with different modalities in the data. Classical regressors usually fulfill only a subset of these properties. In this work, we extend seminal work on Bayesian nonparametric mixtures and derive an efficient variational Bayes inference technique for infinite mixtures of probabilistic local polynomial models with well-calibrated certainty quantification. We highlight the model's power in combining data-driven complexity adaptation, fast prediction, and the ability to deal with discontinuous functions and heteroscedastic noise. We benchmark this technique on a range of large real-world inverse dynamics datasets, showing that the infinite mixture formulation is competitive with classical Local Learning methods and regularizes model complexity by adapting the number of components based on data and without relying on heuristics. Moreover, to showcase the practicality of the approach, we use the learned models for online inverse dynamics control of a Barrett-WAM manipulator, significantly improving the trajectory tracking performance.
\end{abstract}

\begin{IEEEkeywords}
 Hierarchical Local Regression, Inverse Dynamics Control, Fully Generative Models, Dirichlet Process Mixtures.
\end{IEEEkeywords}


\vspace{-0.5cm}
\section{Introduction}
    Principled data-driven, adaptive and incremental learning is a desirable property in domains in which datasets are dynamic and accumulate slowly over time. For example, robots have to build models of their dynamics and the environment as they interact with the world. Moreover, these models have to be computationally efficient during both the learning and evaluation process. In the case of general-purpose robots, these models have also be able to incorporate different modalities of continuous and discrete stochastic random variables and possibly incorporate heteroscedastic noise \cite{todorov2005stochastic, buechler2019control}. Predominant and successful regression techniques, such as Gaussian Process Regression (GPR) \cite{Rasmussen2005Gaussian}, Artificial Neural Networks (ANN) \cite{Goodfellow2016deep}, and Local Regression (LR) \cite{Wasserman2006All}, have a mixed set of properties that are useful in different scenarios. 
    \begin{figure}
        \centering
        \includegraphics[width=0.9\columnwidth]{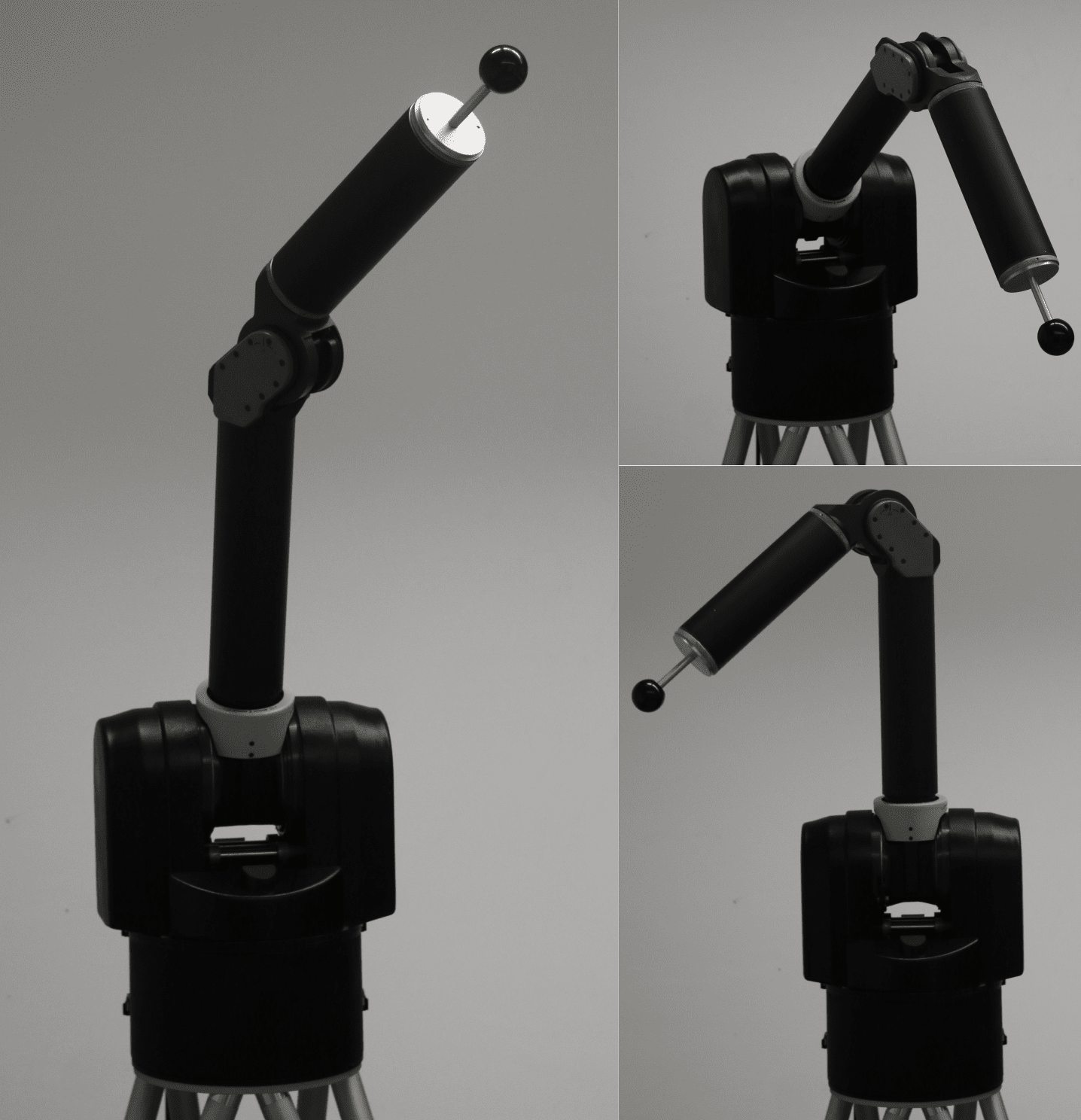}
        \caption{The Barrett-WAM is a challenging cable-driven robot with many unmodelled static and dynamic nonlinearities. Our data-driven inverse dynamics learning technique improves trajectory tracking significantly while offering principled complexity regularization.}
        \label{fig:real_barrett}
        \vspace{-0.55cm}
    \end{figure}
    Gaussian Process Regression offers a principled Bayesian treatment that enables continual and incremental learning, although the $\textit{vanilla}$ formulation of GPR \cite{Rasmussen2005Gaussian} suffered from many drawbacks that have been successfully addressed by recent research, such as the smoothness assumption \cite{Calandra2016Manifold, wilson2016deep, salimbeni2017doubly}, scaling to large datasets \cite{herbrich2003fast, titsias2009variational, bauer2016understanding, matthews2017scalable} and difficulties modeling heteroscedastic noise \cite{le2005heteroscedastic, Kersting2007Most, liu2020large}.
    \begin{figure*}[th]
	    \centering
    	\begin{minipage}{0.49\textwidth}
    		\includegraphics[width=0.95\textwidth]{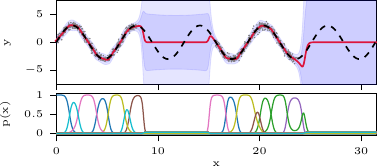}
    	\end{minipage}
    	\hspace{0.1cm}
    	\begin{minipage}{0.46\textwidth}
    		\includegraphics[width=0.95\textwidth]{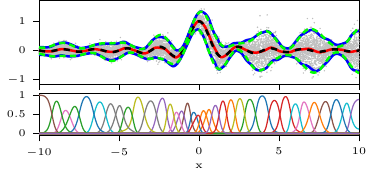}
    	\end{minipage}
    	\vspace{-0.15cm}
    	\caption{Two toy datasets learned with Infinite Local Regression (ILR). The bottom plots show the activation of the local regression models over the input space. \textbf{Left}, an example of how this technique deals with out-of-distribution uncertainty. Mean prediction (red) on training data of a Sine function (grey dots) and the true mean function (black dashed). The shades of the blue area represent a predictive uncertainty of one and two standard deviations. In areas lacking training data the predictive uncertainty of ILR is large, the mean prediction falls back to the prior. \textbf{Right}, the ability of ILR to capture input-dependent noise is depicted. Mean prediction (red) on training data of a Sinc function (grey dots) and the true mean function (black dashed) corrupted by $\pm 2$-$\sigma(x)$ (dashed green). The blue lines represent the $2$-$\sigma$ confidence interval of the mean prediction. ILR is able to learn a strongly input-dependent noise function.}
    	\label{fig:sine_sinc}
        \vspace{-0.5cm}
    \end{figure*}
    Artificial Neural Networks, on the other hand, have proven themselves as very powerful easy-to-train universal approximators. They are however still susceptible to over-parameterization \cite{frankle2018lottery} and catastrophic forgetting \cite{mccloskey1989catastrophic}. Moreover, although major progress on the front of Bayesian Neural Networks (BNN) has been made \cite{Neal1994Bayesian, lakshminarayanan2017simple, khan2018fast}, new evidence suggests that issues regarding the accuracy of uncertainty quantification still need to be tackled \cite{wenzel2020good, foong2020expressiveness}.
    Finally, Local Regression methods have had great success, particularly in the domain of robotics and control, precisely because of their flexibility, ability to model hard nonlinearities and heteroscedastic noise, and the possibility of incorporating new data online. Two categories of LR exist \cite{Ting2010Locally}, \textit{lazy} learners, that maintain all seen data points in memory \cite{Atkeson1997LocallyControl}, and \textit{memoryless} learners that compress data by relying on basis functions in the input space and fitting and storing local parameterized models \cite{Nelles1996Basis, Schaal2002Scalable, Vijayakumar2005Incremental}. However, these methods are often difficult to tune as they possess many hyperparameters. A limited attempt to a Bayesian treatment is made in \cite{ting2009bayesian} to alleviate the need to tune the basis functions by constructing local nonparametric kernels and placing Gamma priors on the kernel width. This approach results in a localized GP formulation that needs to retain the training data in memory, again leading to the computational issues of $\textit{vanilla}$ Gaussian Process Regression. A further Bayesian generalization of LR is suggested in \cite{Meier2014incremental}, in which the authors couple the local models via the loss function, thus reinforcing global coordination, and treat the local models in a Gaussian Regression framework. Finally, an extension of LR to localized Gaussian Process Regression for robotics is presented in \cite{nguyen2008local}. Nonetheless, all mentioned approaches still rely on heuristics for pruning local models and fall short of formulating a full generative model in both the input and target space.
    
    It is essential to distinguish between LR methods and so-called Mixture of Experts (MoE) techniques \cite{Jacobs1991Adaptive, Jordan1994Hierarchical, Rasmussen2002Infinite}. The distinction lies in the fact that MoEs rely on discriminative, explicitly input-dependent gating that divides the input space between \textit{competing} experts. In contrast, LR takes a basis-function approach that resembles generative modeling of the input and allows \textit{cooperation} between all experts during training, enabling each model to influence the global result.

    Following this introduction, it is our opinion that Local Regression with a proper fully Bayesian generative treatment has the potential to serve as an efficient general-purpose probabilistic function approximator with well-calibrated uncertainty estimates and to drive many low-level applications in control and robotics that favor fast models and do not require deep representations. In the following, we set out to present a fully probabilistic graphical mixture model of Local Regression that maintains proper priors on all quantities. We start by presenting what we refer to as the Bayesian \textit{finite mixture} that implicitly assumes a finite number of local regression models. This representation can then be extended to become an \textit{infinite mixture} by relying on the paradigm of Bayesian Nonparametrics (BNP) \cite{hjort2010bayesian} and ultimately results in a formulation related to all aforementioned local methods that alleviates the need for any heuristics. For learning these models, we derive a Variational-Bayes Expectation-Maximization (VBEM) scheme to efficiently infer the posterior parameters \cite{Beal2003Variational} and overcome the need for computationally heavy sampling methods. We benchmark this model on a range of toy tasks to highlight their strengths, as well as on real-world datasets for learning the inverse dynamics of robotic manipulators. Most importantly, we finally use the learned model to perform inverse dynamics control on a real Barrett-WAM manipulator, Figure~\ref{fig:real_barrett}.

    The proposed approach is based on seminal work in the area of Bayesian Nonparametrics. We reference influential Markov chain Monte Carlo (MCMC) sampling techniques for Bayesian Nonparametric density estimation \cite{Escobar1995Bayesian, Neal2000Markov, ishwaran2001gibbs, Rasmussen1999The}, which developed the first seeds of Bayesian Inference for Dirichlet Processes \cite{ferguson1973bayesian} under the Pólya-Urn sampling scheme \cite{blackwell1973ferguson}. Moreover, similar infinite mixture regression models have been proposed in \cite{Mueller1996Bayesian, Shababa2009Nonlinear, Hannah2011Dirichlet, meeds2006alternative, gadd2020enriched}, albeit relying on expensive Gibbs sampling techniques and mostly within a limited application scope. Furthermore, we build on fundamental work from Variational Inference (VI) \cite{attias2000variational}, that presented the first VI algorithm for the Bayesian Finite Gaussian Mixture Model and on the contribution of \cite{BleiVariational2006}, that developed the first VI technique for Infinite Gaussian Mixtures with stick-breaking priors, by relying on a truncated variational posterior representation.


\vspace{-0.15cm}
\section{Preliminaries}
\label{sec:prelim}
    In this section we introduce some concepts related to our approach such as Bayesian Linear Regression, Bayesian Mixture Models the and Dirichlet Process.
    
    \textbf{Bayesian Linear Regression.} We start by discussing the Bayesian treatment of a single component of a fully Bayesian Local Regression model, namely Bayesian Linear Regression \cite{Minka2000Bayesian}. The conditional data model takes a feature vector $\vec{x} \in \mathcal{R}^{m}$ as a random input variable and returns a response random variable $\vec{y} \in \mathcal{R}^{d}$ according to a linear mapping $\mat{A}: \mathcal{R}^{m} \to \mathcal{R}^{d}$ and additive zero-mean noise $\vec{e}$ with a precision matrix $\mat{V}$, $\vec{y} = \mat{A} \vec{x} + \vec{e}, \vec{e} \sim \op{N}(\vec{0}, \mat{V}^{-1})$.
    For a full Bayesian treatment we consider the parameters of this model to be random variables on which we place proper conjugate priors. In this case we place a Matrix-Normal-Wishart prior $\op{MN}(\mat{A} | \mat{M}, \mat{V}^{-1}, \mat{K}^{-1}) \op{W}(\mat{V} | \mat{P}, \eta)$ on the matrix $\mat{A}$ and precision $\mat{V}$,  where $\mat{M}$, the mean of $\mat{A}$, is a $d \times m$ matrix and $\mat{V}$ and $\mat{K}$ are $d \times d$ and $m \times m$ that serve as row and column precision matrices of $\mat{A}$, respectively. The parameters of the Wishart distribution are the $d \times d$ positive definite scale matrix $\mat{P}$ and the degrees of freedom $\eta$. Given the conjugate nature of the chosen priors, the posterior of $p(\mat{A}, \mat{V} | \mathcal{D})$ is also a Matrix-Normal-Wishart distribution, whose parameters can be computed in closed-form, conditioned on a dataset $\mathcal{D}=\{(\vec{x}_{1}, \vec{y}_{1}), \ldots, (\vec{x}_{N}, \vec{y}_{N}) \}$ of $N$ independent and identically distributed data pairs \cite{Minka2000Bayesian}.

    \begin{figure}[t!]
        \centering
	    \includegraphics[width=0.80\columnwidth]{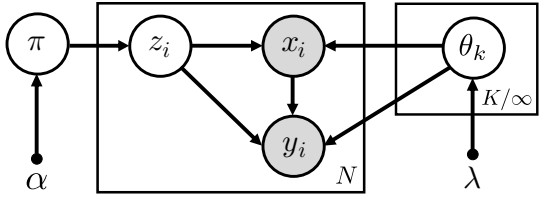}
    	\caption{A unified plate notation for finite and infinite mixtures of Bayesian local regression models. Assuming linear Gaussian mixtures, the parameters $\vec{\theta}_k$ are sampled from the base measure $\op{H}$ with hyperparameters $\lambda$. They contain the means $\vec{\mu}_{k}$ and precision matrices $\mat{\Sigma}_{k}$, drawn from a Normal-Wishart prior, as well as the regression matrices $\mat{A}_{k}$ and output precision matrices $\mat{V}_{k}$, sampled from a Matrix-Normal-Wishart, for every component $k$. The latent variables $z_{i}$ assign every $\vec{x}_{i}$ to a component and are drawn from a categorical distribution parameterized by $\vec{\pi}$. The mixture weights $\vec{\pi}$ are either generated by a finite Dirichlet prior or a stick-breaking process with a concentration parameter $\alpha$. Figures adapted from \cite{Murphy2012Machine}.}
    	\label{fig:plates}
    	\vspace{-0.55cm}
    \end{figure}

    \textbf{Bayesian Finite Mixture Models.} Gaussian mixture models (GMM) are hierarchical latent variable models with universal approximation capabilities for arbitrary continuous densities. This insight is of central interest, when considered later in the context of density estimation for local linear models, that are themselves universal nonlinear function approximators \cite{Wasserman2006All}. A finite $K$-component Gaussian mixture of a random variable $\vec{x}$ is a weighted linear combination of densities
    $ \! p\left(\vec{x} | \vec{\theta} \right) \! = \! \sum_{k=1}^{K} p\left(z| \vec{\pi} \right) p\left(\vec{x} |  \vec{\theta}_{k} \right)  \! = \! \sum_{k=1}^{K} \pi_{k} \op{N}\left(\vec{x} | \vec{\mu}_{k}, \mat{\Sigma}^{-1}_{k }\right) $,
    with $K$ unique mean vectors $\vec{\mu}_{k}$ and precision $\mat{\Sigma}_{k}$. The latent quantity $z \in\{1, \ldots, K\}$ is a discrete random variable, distributed according to a Categorical distribution $p\left(z_{i} \right)=\op{Cat}(\vec{\pi})$, subject to the mixing weights $\vec{\pi} = \{\pi_{1}, \ldots, \pi_{K}\}$, satisfying $0 \leq \pi_{k} \leq 1$ and $\sum_{k=1}^{K} \pi_{k}=1$.
    
    The Bayesian extension \cite{attias2000variational} of this model introduces a conjugate Normal-Wishart prior $\op{H}(\lambda)$ on the means and precision matrices $\vec{\mu}_{k}, \mat{\Sigma}_{k}$, where $\lambda$ is the hyperparameter vector. Furthermore, a conjugate Dirichlet prior, with an $\alpha$ concentration parameter, is placed on the mixing weights $\vec{\pi} \sim \op{Dir}\left((\alpha / K) \vec{1}_{K} \right)$. This Bayesian perspective has proven very effective in regularizing the shortcomings of GMMs, by allowing superfluous components to fall back to their priors, instead of severely over-fitting to small clusters, thus consequently inducing a sparsity effect over $K$ \cite{beal2006variational}.

    \textbf{Dirichlet Process and Stick-Breaking.} A Dirichlet process is a distribution over probability measures $\op{G}$. We write $\op{G} \sim \op{DP}(\alpha, \op{H})$, where $\alpha$ is the \textit{concentration parameter} and $\op{H}$ is the \textit{base measure} \cite{Murphy2012Machine, Teh2010Dirichlet}.
    Intuitively, a Dirichlet process is a distribution over distributions, which means that each draw $\op{G}$ is itself a distribution. The draws from a DP are discrete distributions. The base distribution $\op{H}$ is the mean of the DP and the concentration parameter $\alpha$ can be interpreted as an inverse variance. The larger $\alpha$, the smaller the variance and the DP will concentrate more of its mass around the mean $\op{H}$.
    
    We will rely on the \textit{stick-breaking construction} \cite{Sethuraman1994AConstructive} of a DP as an algorithmic realization. Stick-breaking delivers an infinite sequence of mixture weights $\pi_{k}$ of an infinite mixture model from the probabilistic process $\pi_{k} = v_{k} \prod_{l=1}^{k-1}\left(1-v_{l} \right)$, where $v_{k} \sim \op{Beta}(1, \alpha)$.
    This process is sometimes denoted as $\vec{\pi} \sim \op{GEM}(\alpha)$ \cite{Murphy2012Machine}. The stick-breaking procedure describes how the random variables $v_k$, representing stick lengths, are drawn from a Beta distribution and combined to obtain the mixture weights $\pi_k$. If the concentration parameter $\alpha$ increases, the magnitude of the mixing weights $\pi_k$ decreases on average and the number of probable active components increases.
    This representation of DPs can be used to replace the priors placed on the Finite Gaussian Mixture Model \cite{BleiVariational2006}. In such a setting the sampled measures $\op{G} \sim \op{DP}(\alpha, \op{H(\lambda)})$ is a draw of an unbounded number of parameters $\vec{\theta}_{k} \sim \op{H(\lambda)}$ for an infinite number of clusters, associated with an infinite number of weights $\pi_{k}$ generated by the stick-breaking process. The clustering effect comes into play due to the discrete nature of the DP. Eventually, the same parameters will be sampled over and over, forcing the associated data points to cluster.


\vspace{-0.15cm}
\section{Variational Bayes for Infinite Mixture of Local Regressors}
\label{sec:ilr}

    Using the previously presented concepts of Bayesian Linear Regression, Bayesian Linear Models and the Dirichlet Process, we now construct a fully Bayesian Infinite Mixture of Local Regressors (ILR). Our approach to solving the regression task is mainly a Bayesian joint density estimation task. Our aim is to find the posterior over all regression parameters $p(. | \mathcal{D})$ and use the predictive marginal at prediction time.  We use the generative model as depicted in Figure~\ref{fig:plates}
    \begin{align*}
    	\vec{\pi} &\sim \op{GEM}(\alpha), \quad z_{i} \sim \op{Cat}(\vec{\pi}), \quad \vec{\theta}_{k} \sim \op{H}(\lambda), \\
    	\vec{x}_{i} &\sim \op{N}(\vec{\mu}_{z_{i}}, \mat{\Sigma}^{-1}_{z_{i}}), \quad \vec{y}_{i} \sim \op{N}(\mat{A}_{z_{i}} \vec{x}_{i}, \mat{V}^{-1}_{z_{i}}),
    \end{align*}
    where $\vec{\theta}_{k} = \{ \vec{\mu}_{k}, \mat{\Sigma}_{k}, \mat{A}_{k}, \mat{V}_{k} \}$. Notice here that the defined densities over the input space $\mat{X}$ naturally play the role of the basis functions or so-called receptive fields as in the \textit{Receptive Field Weighted Regression} \cite{Schaal2002Scalable} and \textit{Locally Weighted Projected Regression} \cite{Vijayakumar2005Incremental} algorithms.
    
    In the following, we will tackle the problem of inferring the posterior $p\left(\vec{z}, \vec{v}, \vec{\mu}, \mat{\Sigma}, \mat{A}, \mat{V}\right)$, where $\vec{v}$ are the lengths generated during the stick-breaking process $\op{GEM}(\alpha)$. We opt here for a Variational Bayes approach because VI tend to be more computationally efficient in comparison to MCMC methods, especially considering the high dimensional posterior space in terms of the number of components and parameters per component.

    \textbf{Joint Likelihood Function.} For the general case of multivariate regression with fully populated covariance matrices we use the following likelihood function
    \begin{align*}
    	& p( \mat{Y}, \mat{X}  | \vec{z}, \vec{\mu}, \mat{\Sigma}, \mat{A}, \mat{V})  \\                     
        & = \prod_{n=1}^{N} \prod_{k=1}^{K} \op{N}\left(\vec{y}_{n} | \mat{A}_{k}\vec{x}_{n}, \mat{V}_{k}^{-1}\right)^{z_{n k}} \op{N}\left(\vec{x}_{n} | \vec{\mu}_{k}, \mat{\Sigma}_{k}^{-1}\right)^{z_{n k}},
    \end{align*}
    where the dimensions of all quantities follow the notation of Bayesian Linear Regression.
    
    \textbf{Infinite Conjugate Prior.} We assume the factorized conjugate infinite mixture prior $p\left(\vec{v}, \vec{\mu}, \mat{\Sigma}, \mat{A}, \mat{V}\right)=$
    \begin{align*}
    	 & = p\left(\vec{v}\right) p\left(\mat{A}| \mat{V} \right) p\left(\mat{V} \right) p\left(\vec{\mu}| \mat{\Sigma} \right) p\left(\mat{\Sigma} \right)  \\
    	 & = \prod_{k=1}^{\infty} \op{Beta}\left(v_k | \gamma_{0,1}, \gamma_{0,2 }\right) \prod_{k=1}^{\infty} \op{M N}\left(\mat{A}_{k} | \mat{M}_0, \mat{V}_{k}^{-1}, \mat{K}^{-1}_0 \right)  \\
    	 & ~ \hphantom{=} \op{W} \! \left(\mat{V}_{k} | \mat{P}_{0}, \eta_{0} \right) \! \prod_{k=1}^{\infty} \! \op{N}\left(\vec{\mu}_{k} | \mat{M}_{0}, \left(\lambda_{0} \mat{\Sigma}_{k} \right)^{-1} \right) \! \op{W} \! \left(\mat{\Sigma}_{k} | \mat{L}_{0}, \nu_{0} \right).
    \end{align*}
    This prior introduces a Normal-Wishart distribution on the cluster means $\vec{\mu}_{k}$ and precision matrices $\mat{\Sigma}_{k}$, while a Matrix-Normal-Wishart prior is placed on the regression coefficients $\mat{A}_k$ and the precision matrices $\mat{V}_k$.
    The parameter $\pi_k$ in the Categorical distribution is substituted by the expression of the stick-breaking process $\pi_{k}(\vec{v})=v_{k} \prod_{j=1}^{k-1}\left(1-v_{j}\right)$. The parameters $\vec{v} = \{v_i, \dots, v_K\}$ are independently Beta distributed. By definition, the first hyperparameter of the Beta distribution $\gamma_{0,1}$ is set to $1$. The second hyperparameter $\gamma_{0,2}$ is called concentration parameter and has a crucial impact on the number of components that are learned from the data.
    
    \textbf{Mean-Field Factorization.} For tractable inference we assume a mean field factorization of the posterior $p\left(\vec{z}, \vec{v}, \vec{\mu}, \mat{\Sigma}, \mat{A}, \mat{V} \lvert \vec{x} \right) \approx \prod_{n=1}^{N} q\left(\vec{z}_{n}\right) q\left(\vec{v}, \vec{\mu}, \mat{\Sigma},\mat{A}, \mat{V} \right)$, which makes the assumptions that latent variables $\vec{z}$ are conditionally independent from the other parameters.
    
    \textbf{Truncated Variational Posterior.} The posterior is a result of the free form optimization of the variational distributions using the
    mean-field equation \cite{Bishop2007Pattern}. We follow \cite{BleiVariational2006} by allowing a truncated form of the posterior, while maintaining an infinite prior. This truncation is considered in contrast to approaches that truncate the Dirichlet process \cite{ishwaran2001gibbs}. During evaluation we choose a very high truncation threshold that is seldom reached. The resulting approximate variational posterior has the following conjugate form $q^{\star}\left(\vec{z}, \vec{v}, \vec{\mu}, \mat{\Sigma},\mat{A}, \mat{V} \right) =$
   	\vspace{-0.025cm}
	\begin{equation}
		\vspace{-0.2cm}
		\begin{aligned}
			= & \prod_{n=1}^{N} \op{Cat} \left(\vec{z}_{n} | \vec{r}_{n} \right) \prod_{k=1}^{K-1}\op{Beta}\left(v_k | \gamma_{k,1}, \gamma_{k,2} \right) \\[0.01ex]
			& \prod_{k=1}^{K} \op{N}\left(\vec{\mu}_{k} | \mat{M}_{k},\left(\lambda_{k} \mat{\Sigma}_{k} \right)^{-1} \right) \op{W}\left(\mat{\Sigma}_{k} | \mat{L}_{k}, \nu_{k} \right) \\[0.01ex]
			& \prod_{k=1}^{K}\op{MN}\left(\mat{A}_{k} | \mat{M}_k, \mat{V}_{k}^{-1}, \mat{K}^{-1}_k \right) \op{W}\left(\mat{V}_{k} | \mat{P}_{k}, \eta_{k} \right). \\[0.05ex]
		\end{aligned}
	\end{equation}

	\vspace{0.025cm}
    \textbf{Variational Expectation Step.} In the E-step the responsibilities are computed, while other variational parameters are  fixed. The responsibilities are variational parameters of the categorical $q^{\star}(\vec{z}) =\prod_{n=1}^{N} \prod_{k=1}^{K} r_{n k}^{z_{n k}}$ and are calculated as
    \begin{align*}
    	 r_{n k} \propto & \tilde{\mat{V}}_{k}^{\frac{1}{2}} \tilde{\mat{\Sigma}}_{k}^{\frac{1}{2}} \exp \left[-\frac{m}{2 \lambda_{k}} + \mathbb{E}_{q}\left[\log v_{k} \right]\right. \\
    	 & -\frac{\nu_{k}}{2} \! \left(\vec{x}_{n}-\mat{M}_{k }\right)^{\top} \! \mat{L}_{k} \! \left(\vec{x}_{n}-\mat{M}_{k} \right) \! -  \! \frac{1}{2} \! \op{Tr}\left(\mat{K}_{k}^{-1}\vec{x}_n  \vec{x}_{n}^{\top} \right) \\
    	 & - \frac{\eta_k }{2} \left( \vec{y}_n - \mat{M}_k \vec{x}_{n} \right)^{\top} \mat{P}_k \left( \vec{y}_n - \mat{M}_k \vec{x}_{n} \right) \\
    	 & + \sum_{j=1}^{k-1} \mathbb{E}_{q}\left[\log \left(1-v_{j} \right)\right] \Bigg],
    \end{align*}
    with the help of the following expressions
    \begin{align*}
    	 \mathrm{E}_{q} \left[\log v_{k}\right] & = \psi\left(\gamma_{k, 1}\right)-\psi\left(\gamma_{k, 1} + \gamma_{k, 2} \right), \\
    	 \mathrm{E}_{q} \left[\log \left(1-v_{k} \right)\right] & = \psi\left(\gamma_{k, 2} \right) - \psi\left(\gamma_{k, 1}+\gamma_{k, 2} \right), \\
    	 \log \tilde{\mat{\Sigma}}_{k} & = \sum_{j=1}^{m} \psi\left(\frac{\nu_{k}+1-j}{2}\right)+ \log \left|\mat{\Sigma}_{k} \right|, \\ 
    	 \log \tilde{\mat{V}}_{k} & = \sum_{j=1}^{d} \psi\left(\frac{\eta_{k}+1-j}{2}\right)+\log \left|\mat{V}_{k} \right|,
    \end{align*}
    where $\psi$ is the Digamma function.
    
    \textbf{Variational Maximization Step.} The M-step updates the remaining variational parameters as follows
    \begin{align*}
    	\gamma_{k,1}          & = \sum_{n=1}^{N} r_{n k} + \gamma_{0,1}, \quad \gamma_{k,2} = \sum_{n=1}^{N} \sum_{j=k+1}^{K} r_{n j} + \gamma_{0,2}, \\
    	\lambda_k             & = \lambda_0 + N_k , \quad \mat{M}_k = \frac{1}{\lambda_k} \left( N_k \overline{\vec{x}}_k + \lambda_0 \mat{M}_0 \right), \\
    	v_{k} & =v_{0}+N_{k}, \quad \eta_k = \eta_0 + N_k, \\
    	\! \mat{L}_{k}^{-1} \! & = \! \mat{L}_{0}^{-1} \! + \! N_{k} \mat{S}_{k} \! + \! \frac{\lambda_{0} N_{k}}{\lambda_{k}+ N_{k}}\!\left(\mat{M}_{0}-\overline{\vec{x}}_{k}\right)\!\left(\mat{M}_{0}-\overline{\vec{x}}_{k}\right)^{\top}, \\ 
    	\mat{K}_k & =  \sum_{n=1}^{N} r_{nk} \vec{x}_n \vec{x}_{n}^{\top}  + \mat{K}_{0},\\
    	\! \mat{M}_k           & = \left[ \sum_{n=1}^{N} r_{nk} \vec{y}_{n} \vec{x}_{n}^{\top}  + \mat{M}_{0} \mat{K}_0  \right] \mat{K}_{k}^{-1}, \\
    	\mat{P}_{k}^{-1} \! & = \! \mat{P}_{0}^{-1} \! + \! \mat{M}_0 \mat{K}_0 \mat{M}_{0}^{\top} \! + \! \sum_{n=1}^{N} \! r_{n k} \vec{y}_n \vec{y}_{n}^{\top} \! - \! \mat{M}_k \mat{K}_k \mat{M}_{k}^{\top},
    \end{align*}
    where the data statistics are defined as
    \begin{align*}
    	N_{k}      & = \sum_{n=1}^{N} r_{n k}, \quad \overline{\vec{x}}_{k} = \frac{1}{N_k}\sum_{n=1}^{N} r_{nk} \vec{x}_n, \\
    	\mat{S}_{k} & = \frac{1}{N_{k}} \sum_{n=1}^{N} r_{n k}\left(\vec{x}_{i}-\overline{\vec{x}}_{k} \right) \left(\vec{x}_{i}-\overline{\vec{x}}_{k} \right)^{\top}.
    \end{align*}

    \textbf{Posterior Predictive Distribution.} For predicting the function value $\hat{\vec{y}}$ conditioned on a test query $\hat{\vec{x}}$ we marginalize the likelihood over the posterior parameters $\vec{\theta}$ to get the joint posterior predictive density, which results in a mixture of Student's t-distributions. To make the marginalization tractable, we replace the true posterior by our approximate variational posterior inferred under a training dataset $\mathcal{D}$ \cite{Bishop2007Pattern}.

    \textbf{Computational Complexity.} Given the previous equations, we can deduce that the training-time computational cost is $\mathcal{O}(NK(d+m)^3)$ and can be straightforwardly reduced to $\mathcal{O}(MK(d+m)^3)$ by applying stochastic updates \cite{hoffman2013stochastic}, where $M$ is the batch size. This result shows linear scalability with the data, which is considerably more efficient than simple variants of GPR. The testing-time complexity of a mean prediction is $\mathcal{O}(K(d^3+dm))$ which combines the input membership query and the linear matrix transformation for every model $k$. This computation is, in contrast to GPs, independent of the training data size, hence the advantage of memoryless representations during real-time critical applications.
    

\vspace{-0.1cm}
\section{Empirical Evaluation}
\label{sec:result}

    We evaluate different aspects of the proposed model on a range of tasks. Our goals are \textbf{1)} to highlight some of the advantages of ILR, such as dealing with out-of-distribution predictions, recovering an input-dependent noise function, and the ability to perform Bayesian sequential updates, \textbf{2)} to benchmark the model on high dimensional datasets from real robots, and \textbf{3)} to deploy the model in a real-world scenario to further empirically demonstrate its validity. The source code can be found under \url{https://github.com/hanyas/mimo}.
    
    \textbf{Out-of-distribution Uncertainty.} In Figure~\ref{fig:sine_sinc} (left), we apply ILR to a synthetic Sine dataset with two large gaps. We observe that the predictive uncertainty strongly reflects the lack of training data in these regions. Furthermore, the mean prediction falls back to the prior values quickly. This example highlights the reasonable quantification of uncertainty by the model. Uncertainty is low, where the mean prediction is accurate and very high in regions where the prior dominates. The out-of-distribution behavior of ILR is strongly influenced by the input gating and a query's membership probability.
    
    \textbf{Input-dependent Noise Function.} Following \cite{liu2020large}, we generate data from $y(x)=\op{sinc}(x)+\epsilon, \quad x \in[-10,10]$, where the noise term $\epsilon$ is distributed according to a normal distribution $\op{N}\left(0, \sigma_{\epsilon}^{2}(x) \right)$ with input-dependent standard deviation $\sigma_{\epsilon}(x)=0.05+0.2(1+$ $\sin (2 x)) /\left(1+e^{-0.2 x} \right)$. Figure~\ref{fig:sine_sinc} (right) shows that ILR can approximate the nonlinear and highly heteroscedastic function well. In particular, the input-dependent noise function is recovered in great detail.

    \begin{figure}
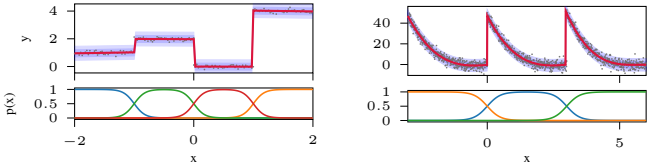

    	\centering
    	\begin{minipage}[b]{0.43\columnwidth}
    		\centering
    		\input{figures/step.tex}
    	\end{minipage}
    	\hspace{0.65cm}
    	\begin{minipage}[b]{0.43\columnwidth}
    		\centering
            \input{figures/step_poly_features.tex}
    	\end{minipage}
    	\vspace{-0.5cm}
    	\caption{\textbf{Discontinuous functions.} The bottom plots show the activation of the regressors over the input space.
    	Mode (red) and $2$-$\sigma$ confidence interval (shaded blue) of the predictive distribution fitted to data (gray dots). \textbf{Left}, the figure shows that ILR can approximate discontinuous data, such as a step function. \textbf{Right}, using a polynomial feature transformation gives the model more flexibility.}
    	\label{fig:step}
    	\vspace{-0.25cm}
    \end{figure}
    
    \textbf{Discontinuous Data and Local Polynomials.} In Figure~\ref{fig:step} (left), a combination of step functions is fitted using the mode of the predictive distribution. By applying a polynomial feature transformation to the input space, more expressive local regressors can be realized. Figure~\ref{fig:step} (right) depicts an example of cubic regressors, which are still linear in the parameters, fitted to data sampled from noisy cubic polynomials.
    
\begin{figure}[t]
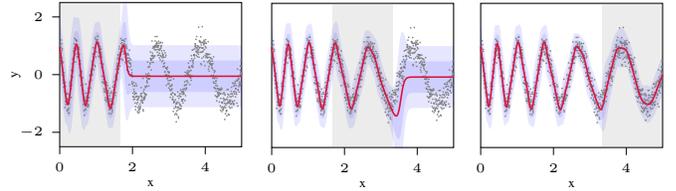

		\centering
		\begin{minipage}{0.35\linewidth}
			\centering
			\input{figures/chirp_0.tex}
		\end{minipage}\hspace{0.2cm}
		\begin{minipage}{0.3\linewidth}
			\vspace{-0.4cm}
			\input{figures/chirp_1.tex}
		\end{minipage}
		\begin{minipage}{0.3\linewidth}
			\vspace{-0.4cm}
			\input{figures/chirp_2.tex}
		\end{minipage}
		\vspace{-0.45cm}
		\caption{\textbf{Bayesian sequential updates.} Mean (red) and $2$-$\sigma$ confidence interval (shaded blue) of the predictive distribution fitted to sequentially arriving data (three batches) from the Chirp dataset (gray dots). For the second and third plot the posterior fitted to the previous batches is used as a prior to perform a Bayesian sequential update. There is no catastrophic forgetting of previously learned knowledge and in regions with no data the prediction falls back to the prior.}
		\label{fig:bayes_seq}
		\vspace{-0.25cm}
	\end{figure}

    \textbf{Bayesian Sequential Updates.} In Figure~\ref{fig:bayes_seq} we demonstrate a sequential learning problem. Data from the Chirp dataset arrives in three batches, and the posterior of the previous batches becomes the prior for the current learning batch. ILR successfully captures the data trend and no significant catastrophic forgetting of previously learned knowledge can be observed. The approximation errors resulting from the mean-field assumption has little influence because the posterior updates are localized in the input domain.
    
	\begin{table}[th]
        \centering
        \renewcommand{\arraystretch}{1}
		\caption{Accuracy on the SARCOS inverse dynamics dataset.}
        \sisetup{scientific-notation=fixed, table-align-uncertainty=true, separate-uncertainty=true}
        \setlength\tabcolsep{3.5pt}
		\begin{tabular}{c | S[fixed-exponent=-1] | S[fixed-exponent=-3] | S} 
		    \toprule[1.1pt]
			        & {MSE}                   & {NMSE}                      & {Models}    \\ \midrule
			ILR     & \num{0.480 \pm 0.030}   & \num{0.00340 \pm 0.0002}    & \num{1700}  \\
			LGR$^{*}$  & \num{8.600 \pm 0.000}   & \num{0.05000 \pm 0.0000}    & \num{7000}   \\
			LWPR    & \num{2.600 \pm 0.030}   & \num{0.01800 \pm 0.0002}    & \num{32000}  \\
			GPR$^{\dagger}$   & \num{0.610 \pm 0.020}   & \num{0.0041  \pm 0.00007}   & {-}  \\
			SGPR	& \num{0.850 \pm 0.003}   & \num{0.006000 \pm 0.000008}	& {-} \\ 
			\bottomrule[1.1pt]
		\end{tabular}
		\label{tab:sota_sarcos}
	\end{table}
	
	\begin{table}[h!]
	    \centering
	    \renewcommand{\arraystretch}{1}
		\caption{Accuracy on the Barrett inverse dynamics dataset.}
        \sisetup{scientific-notation=fixed, table-align-uncertainty=true, separate-uncertainty=true}
        \setlength\tabcolsep{3.5pt}
		\begin{tabular}{c | S[fixed-exponent=-1] | S[fixed-exponent=-3] | S} 
		    \toprule[1.1pt]
			        & {MSE}                   & {NMSE}                      & {Models}    \\ \midrule
			ILR     & \num{0.290 \pm 0.050}   & \num{0.0070 \pm 0.0005}     & \num{1350} \\
			LGR$^{*}$  & \num{0.770 \pm 0.000}   & \num{0.0170 \pm 0.000}      & \num{3270} \\
			LWPR    & \num{1.000 \pm 0.150}   & \num{0.0370 \pm 0.0100}     & \num{2900} \\ 
			GPR     & \num{0.100 \pm 0.0003}  & \num{0.0023 \pm 0.00001}    & {-} \\
			SGPR    & \num{0.180 \pm 0.005}   & \num{0.0063 \pm 0.00002}    &  {-} \\ 
			\bottomrule[1.1pt]
		\end{tabular}
		\vspace{-0.5cm}
		\label{tab:sota_barrett}
	\end{table}
	
    Next we learn the inverse dynamics of real anthropomorphic manipulators governed by
    $\mat{M}(\vec{q}) \ddot{\vec{q}}+\mat{C}(\vec{q}, \dot{\vec{q}})+\mat{G}(\vec{q})+\epsilon(\vec{q}, \dot{\vec{q}}, \ddot{\vec{q}})=\vec{u}$, where $\vec{q}, \dot{\vec{q}}, \ddot{\vec{q}}$ are joint angles, velocities and accelerations and $\vec{u}$ are the torques. $\mat{M}(\vec{q})$ is the inertia matrix, $\mat{C}(\vec{q}, \dot{\vec{q}})$ are the Coriolis and centripetal forces and $\mat{G}(\vec{q})$ is the gravity force. $\epsilon(\vec{q}, \dot{\vec{q}}, \ddot{\vec{q}})$ are general unmodelled nonlinearities such as sticktion/friction and hydraulic and tendon/cable dynamics, we thus take a data-driven approach to learn the mapping $\vec{q}, \dot{\vec{q}}, \ddot{\vec{q}} \rightarrow \vec{u}$. The learned model is then used for model-based control. We use the Mean Squared Error (MSE), Normalized Mean Squared Error (NMSE) and the number of active models as evaluation criteria. These measures cover the prediction accuracy, as well as the complexity of the learned model. We compare to popular (probabilistic) methods such as LGR \cite{Meier2014incremental}, LWPR \cite{Vijayakumar2005Incremental}, GPR \cite{Rasmussen2005Gaussian} and SGPR \cite{titsias2009variational}\footnote{(S)GPR use an RBF-kernel with hyperparameter optimization.}. 
    
    \textbf{SARCOS and Barrett-WAM Datasets.} We benchmark the prediction accuracy of ILR on a relatively high-dimensional dataset collected from a 7-DoF anthropomorphic SARCOS arm \cite{Vijayakumar2005Incremental}. The dataset consists of 44484 training points and 4449 test cases. Overall there are 21 input variables, $\vec{q}, \dot{\vec{q}}, \ddot{\vec{q}}$, mapping to 7 motor torques $\vec{u}$. We also benchmark on a real-world inverse dynamics dataset of a 4-DoF Barrett-WAM manipulator, Figure~\ref{fig:real_barrett}, where we map from a 12-D to a 4-D space. The dataset contains 25000 training pairs and 5000 test cases.
    Table~\ref{tab:sota_sarcos} and Table~\ref{tab:sota_barrett} list the results for both datasets. We report the average MSE, NMSE, and number of active models over all joints. The results are obtained by running five seeds and computing the means and standard deviations for every cell in the table, except for LGR$^{*}$, because of the unreasonable training times achieved while using the authors' code. When evaluating GPR on the SARCOS dataset, we faced GPU-memory constraints and had to limit the dataset to 35000 instead of 44484 training samples.
    The results of this evaluation show that ILR clearly outperforms LWPR and LGR on both datasets, in terms of prediction accuracy and number of used models. On the SARCOS dataset, ILR even outperforms GPR. We attribute this result however, to the fact that GPR had to work with a reduced training set$^{\dagger}$, and acknowledge that GPR is still the gold standard on medium-sized datasets. Finally, the results indicate that ILR is, in general, very competitive with SGPR.
    

    \begin{figure}[t]
    	\centering
\begin{tikzpicture}

\definecolor{color0}{rgb}{0.12156862745098,0.466666666666667,0.705882352941177}
\definecolor{color1}{rgb}{1,0.498039215686275,0.0549019607843137}
\definecolor{color2}{rgb}{0.172549019607843,0.627450980392157,0.172549019607843}
\definecolor{color3}{rgb}{0.83921568627451,0.152941176470588,0.156862745098039}
\definecolor{color4}{rgb}{0.580392156862745,0.403921568627451,0.741176470588235}

\begin{axis}[
log basis y={10},
tick align=outside,
tick pos=left,
grid=both,
minor tick num=3,
x grid style={white!69.0196078431373!black},
xlabel={Batch},
ylabel={NMSE},
xmin=-0.7, xmax=14.7,
xtick style={color=black},
y grid style={white!69.0196078431373!black},
ymin=0.00144678257611342, ymax=0.106772863328083,
ymode=log,
y label style={yshift=-1.5em, xshift=1.5em},
x label style={yshift=0.5em},
height=4cm, width=9cm
]

\addplot [semithick, color0, mark=diamond*, mark size=3, mark options={solid}]
table {%
0 0.0131312119504423
1 0.00513998710983266
2 0.00383082666881862
3 0.0033990267774292
4 0.00292341179999811
5 0.00263507763585391
6 0.00240194716989073
7 0.00232028366502812
8 0.00222128846578296
9 0.00213317798777379
10 0.00204148511152258
11 0.00195878115119563
12 0.0018947602960907
13 0.0018165648939239
14 0.00175919904181832
};
\addplot [semithick, color1, mark=diamond*, mark size=3, mark options={solid}]
table {%
0 0.0878110518438791
1 0.0254066884372742
2 0.0157127897623549
3 0.0124993288938106
4 0.00990174962286861
5 0.00832129206268462
6 0.00753887308341494
7 0.0069280111021065
8 0.00621051299795261
9 0.00560021665384591
10 0.00515233179324981
11 0.0047256976478639
12 0.00426273186255177
13 0.00398470359495318
14 0.00387041745038119
};
\addplot [semithick, color2, mark=diamond*, mark size=3, mark options={solid}]
table {%
0 0.0454678216092707
1 0.0186223873086292
2 0.0106517348953159
3 0.00726022611851029
4 0.0057279025311987
5 0.00518677156522962
6 0.00465771774252888
7 0.00417530995324389
8 0.00374950802068141
9 0.00351238798703091
10 0.00333925914214395
11 0.00314319416427888
12 0.00313539567954757
13 0.00295658332934545
14 0.0027640179163555
};
\addplot [semithick, color3, mark=diamond*, mark size=3, mark options={solid}]
table {%
0 0.0335585220570522
1 0.0168204663963333
2 0.0111190064367559
3 0.00791004448920563
4 0.00619554709237158
5 0.00536391407362902
6 0.00479328039693672
7 0.00435233581256755
8 0.0040341162512989
9 0.00382698565519801
10 0.00361386146780918
11 0.00345671011838333
12 0.00328708880618245
13 0.00317252226168252
14 0.00300220812760887
};
\addplot [semithick, color4, mark=diamond*, mark size=3, mark options={solid}]
table {%
0 0.020283665147216
1 0.00664636545304875
2 0.00519629022299606
3 0.00481096972004358
4 0.00427615182659968
5 0.00389282167699856
6 0.00359952062329616
7 0.00339472471175262
8 0.00336421082321336
9 0.00323978477084708
10 0.00311499397072179
11 0.00316010303138192
12 0.00311072026457559
13 0.00301454141700375
14 0.00302835424824366
};
\end{axis}

\end{tikzpicture}
    	\caption{\textbf{8-Shaped Trajectory Learning.} Bayesian sequential updates on the 8-shaped trajectory dataset collected from a Barrett-WAM. For 5 different random seeds, we plot the NMSE on accumulated data over the number of batches. The NMSE consistently improves with new data and no catastrophic forgetting is observed.}
    	\label{fig:eight_bayesian}
    	\vspace{-0.25cm}
    \end{figure}
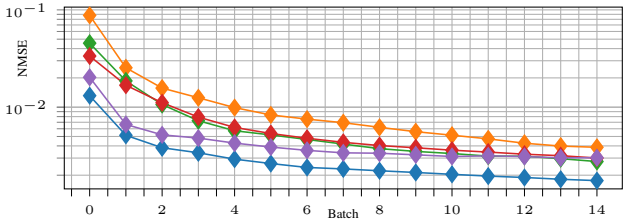

    \begin{figure}[t]
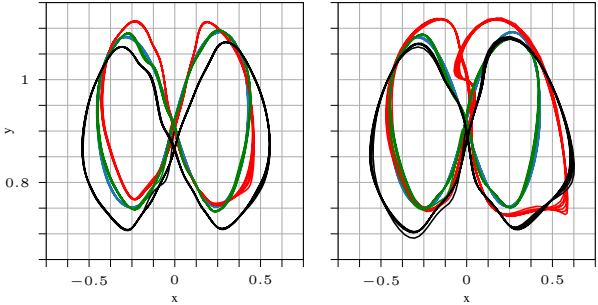

    	\centering
    	\begin{minipage}[b]{0.43\columnwidth}
    		\centering
    		\input{figures/eight_0275.tex}
    	\end{minipage}\hspace{0.5cm}
    	\begin{minipage}[b]{0.4\columnwidth}
    		\centering
            \input{figures/eight_0425.tex}
    	\end{minipage}
    	\vspace{-0.5cm}
    	\caption{\textbf{8-Shaped Trajectory Tracking.} Qualitative results of tracking two 8-shaped test trajectories with different execution speeds (blue) on a Barrett-WAM, while applying a low-gain PD-controller (black), low-gain PD + feed-forward torques from an analytical model (red), low-gain PD + feed-forward torques from ILR (green).}
    	\label{fig:eight_tracking}
    	\vspace{-0.65cm}
    \end{figure}
    
    \textbf{Real Inverse Dynamics Control.} We demonstrate the validity of the learned dynamics captured by ILR by using the learned model in an online trajectory tracking scenario with inverse dynamics control on the Barrett-WAM, see Figure~\ref{fig:real_barrett}. In this experiment, we learn two separate models for two different tasks. The \textbf{first task} requires tracking an 8-shaped desired trajectory in the $xy$-plane of the end-effector. We collect 30000 training samples (roughly 1 minute) consisting of multiple trajectories with different velocity profiles. We perform learning with Bayesian sequential updates over 15 localized batches for multiple seeds. Figure~\ref{fig:eight_bayesian} depicts the progression of the learning process. The NMSE over accumulated data improves consistently, despite the mean-field posterior approximation. We then select the best model and perform online model-based control to track held-out test trajectories with unseen velocity profiles. ILR provides feed-forward torques while supported by a low-gain PD-controller. We compare the tracking precision of ILR to that of an analytical dynamics model accompanied by the same low-gain PD-controller and to the tracking precision of a "model-free" PD-controller. Figure~\ref{fig:eight_tracking} shows a qualitative comparison of the different controllers on two test trajectories. For the \textbf{second task}, we construct a similar scenario, albeit we learn a model that covers a larger region of the state-action space and compute quantitative precision benchmarks. We generate a larger real-world Barrett dataset consisting of $150000$ training examples (roughly 5 minutes). The movements are sinusoidal joint-space trajectories with slow and fast velocity profiles. We repeat the process of the previous task and run ILR on held-out test trajectories with the same low-gain PD-controller and compare with the analytical model and "model-free" PD. As benchmarking criteria, we evaluate the MSE w.r.t. the desired trajectory and the mean torque contributed by the low-gain PD-controller to the overall control signal. The rationale is as follows; A good inverse dynamics model will consistently produce a low MSE while not relying on the PD-controller's assistance in the background. Table~\ref{tab:invdyn} shows the benchmark quantities for 3 test trajectories. The results indicate that ILR outperforms both control strategies and achieves good tracking with little contribution from the PD-controller. During both tasks, we are able to consistently achieve a prediction frequency of 2000~Hz, while the Barrett requires 500~Hz.

    \begin{table}[t!]
    	\caption{Tracking error and torque contributed by the PD-controller during the Barrett-WAM real robot task.}
    	\vspace{-0.25cm}
    	\footnotesize
    	\begin{center}
    	\setlength\tabcolsep{6pt}
    	\begin{tabular}{c | c | c | c | c } 
    	    \toprule[1.1pt]
    	        &   &   PD   &  Analytic+PD   & ILR+PD \\ \midrule
    	     \multirow{2}{*}{T1} & MSE       & $2.33 \times 10^{-2}$ & $2.16 \times 10^{-2}$ & $\bm{1.03 \times 10^{-3}}$\\
    	                         & PD-Torque & $8.25 \times 10^{0}$ & $7.12 \times 10^{0}$ & $\bm{1.40 \times 10^{0}}$\\
    	                         \midrule       
    	     \multirow{2}{*}{T2} & MSE       & $2.60 \times 10^{-2}$ & $2.55 \times 10^{-2}$ & $\bm{9.17 \times 10^{-4}}$\\
    	                         & PD-Torque & $8.71 \times 10^{0}$ & $7.41 \times 10^{0}$ & $\bm{1.33 \times 10^{0}}$\\
    	                         \midrule
    	     \multirow{2}{*}{T3} & MSE       & $2.94 \times 10^{-2}$ & $3.08 \times 10^{-2}$ & $\bm{8.96 \times 10^{-4}}$\\
    	                         & PD-Torque & $9.38 \times 10^{0}$ & $8.06 \times 10^{0}$ & $\bm{1.38 \times 10^{0}}$\\
    		\bottomrule[1.1pt]
    	\end{tabular}
    	\end{center}
    	\label{tab:invdyn}
    	\vspace{-0.75cm}
    \end{table}

\vspace{-0.1cm}
\section{Conclusion}
\label{sec:conclusion}

    We presented and successfully applied a computationally efficient Variational Bayes technique for learning inverse dynamics, which is based on the principles of infinite mixtures and Bayesian Nonparametrics. We situated this approach as the next iteration in a large family of local regression techniques such as RFWR, LWPR, and LGR. We showed that placing Dirichlet Process priors on Bayesian mixtures of local regression models can regularize model complexity with little loss in performance and without relying on heuristics. Empirical evaluations indicate that the model offers well-calibrated uncertainty quantification, outperforms LWPR and LGR, and is competitive with Sparse GPR. Moreover, we have confirmed the practicality this approach for online inverse dynamics control on a Barrett-WAM. Further research will consider higher model compression by constructing hierarchical models that enable parameter sharing between local regressors.
    

\clearpage

\bibliographystyle{IEEEtran}
\bibliography{references}

\end{document}